\documentclass[12pt]{spieman}  % 12pt font required by SPIE;
\usepackage{amsmath,amsfonts,amssymb}
\usepackage{graphicx}
\usepackage{setspace}
\usepackage{multirow}
\usepackage{subfigure}

\title{Transferability of Deep Learning Algorithms for Malignancy Detection in Confocal Laser Endomicroscopy Images from Different Anatomical Locations of the Upper Gastrointestinal Tract}

\author[a*]{Marc Aubreville}
\author[b]{Miguel Goncalves}
\author[c,d]{Christian Knipfer}
\author[e,d]{Nicolai Oetter}
\author[f]{Helmut Neumann}
\author[e,d]{Florian Stelzle}
\author[g]{Christopher Bohr}
\author[a,d]{Andreas Maier}
\affil[a]{Pattern Recognition Lab, Computer Science, Friedrich-Alexander-Universit{\"a}t Erlangen-N{\"u}rnberg, Germany}
\affil[b]{Department of Otorhinolaryngology, Head and Neck Surgery, University Hospital Erlangen, Friedrich-Alexander-Universit{\"a}t Erlangen-N{\"u}rnberg, Germany}
\affil[c]{Department of Oral and Maxillofacial Surgery, University Medical Center Hamburg-Eppendorf, Germany}
\affil[d]{Erlangen Graduate School in Advanced Optical Technologies (SAOT), Friedrich-Alexander-Universit{\"a}t Erlangen-N{\"u}rnberg, Germany}
\affil[e]{Department of Oral and Maxillofacial Surgery, University Hospital Erlangen, Friedrich-Alexander-Universit{\"a}t Erlangen-N{\"u}rnberg, Germany}
\affil[f]{First Department of Internal Medicine, University Medical Center Mainz, Johannes Gutenberg-Universit{\"a}t Mainz, Germany}
\affil[g]{Department of Otorhinolaryngology, Head and Neck Surgery, Universit{\"a}t Regensburg, University Hospital, Regensburg, Germany}

\begin{document} 
\maketitle

\begin{abstract}
Squamous Cell Carcinoma (SCC) is the most common cancer type of the epithelium and is often detected at a late stage. Besides invasive diagnosis of SCC by means of biopsy and histo-pathologic assessment, Confocal Laser Endomicroscopy (CLE) has emerged as noninvasive method that was successfully used to diagnose SCC in vivo. For interpretation of CLE images, however, extensive training is required, which limits its applicability and use in clinical practice of the method. 

To aid diagnosis of SCC in a broader scope, automatic detection methods have been proposed. This work compares two methods with regard to their applicability in a transfer learning sense, i.e. training on one tissue type (from one clinical team) and applying the learnt classification system to another entity (different anatomy, different clinical team). Besides a previously proposed, patch-based method based on convolutional neural networks, a novel classification method on image level (based on a pre-trained Inception V.3 network with dedicated preprocessing and interpretation of class activation maps) is proposed and evaluated. 

The newly presented approach improves recognition performance, yielding accuracies of $91.63\%$ on the first data set (oral cavity) and $92.63\%$ on a joint data set. The generalization from oral cavity to the second data set (vocal folds) lead to similar area-under-the-ROC curve values than a direct training on the vocal folds data set, indicating good generalization.  
\end{abstract}

% Include a list of up to six keywords after the abstract

% Include email contact information for corresponding author
%{\noindent \footnotesize\textbf{*}Marc Aubreville,  \linkable{marc.aubreville@fau.de} }

\begin{spacing}{1}   % use double spacing for rest of manuscript

\textbf{Erratum}: In earlier versions of this paper, the count of CLE image sequences for the vocal folds was inadequately reported to be 73. In reality, it was 47.  

\section{Introduction}
\noindent 
Squamous cell carcinoma (SCC) is the most common kind of cancer of the epithelium, which accounts for over 90 percent of all cancer types in the oral cavity and pharynx \cite{Forastiere:2009bw}, as well as for almost the totality of malignancies in the larynx \cite{Krebsregisterdaten:2017tj}. For SCC, the incidence rates are higher for men in their 6th and 7th decade, and long-term tabacco and alcohol consumption are regarded as the most important risk factors \cite{Maier:uh,Westra:2015dt}. 
Only one third of the patients with head and neck cancer is diagnosed in an early tumor stadium (T1), and thus treatment options are reduced and the need for more radical surgical treatment is often increased \cite{Muto:2004hy}.

The gold standard of diagnosis is the invasive biopsy of the epithelial tissue with subsequent histopathological assessment. The biopsies and (part) resections of mutated tissue provide  information about surgical margins, i.e. healthy tissue is biopsied to demonstrate the disease has not spread beyond the resection area. However, due to the invasiveness a limitation in the sample size and quantity can hinder the finding of accurate resection margins and limit monitoring of these lesions. The resection volume is correlated with the severity of functional disorders (e.g. concerning swallowing, speech, voice). Thus, for surgical removal of SCC, the resection volume should be as low as reasonably possible while completely removing the tumor. A non- or minimally invasive in-vivo characterization of microstructures would be a significant asset for the early diagnosis of SCC while at the same time reducing the aforementioned risks. Further, it could significantly help to improve tumor follow-up monitoring of possible local recurrence, reducing the risk accompanying unnecessary biopsies.

One non-invasive method that has successfully been used to differentiate micro-cellular structures is Confocal Laser Endomicroscopy (CLE) \cite{Neumann:2010hb,Oetter:2016cp,Goncalves:2017wn}. For this fluorescence imaging method, laser light in the cyan color spectrum is emitted by a laser diode and directed towards the tissue under investigation by means of a fiber bundle, which is typically inserted into cavities of the body using the accessory channel of an endoscope \cite{Chauhan:2014im}. The resolution of CLE is high, providing magnifications of up to 1000x \cite{Oetter:2016cp} and enabling sub-cellular imaging in real-time. A contrast agent (typically fluorescein) is applied intravenously prior to the examination in order to stain the intercellular space and hence outline the cell borders. 

CLE is being successfully used in diagnostics of the intestine in clinical practice \cite{Neumann:2010hb}. It was reported, that probe-based CLE (pCLE) can also successfully be used for assessment of malignancy of SCC of the vocal folds \cite{Goncalves:wn,Goncalves:2017wn} as well as in epithelium of the oral cavity \cite{Oetter:2016cp}.  However, it is also known that interpretation of CLE images requires significant training \cite{Neumann:2011tc}. An automatic interpretation and classification of CLE images could thus help to spread this non-invasive method in screening more widely, and also improve identifying the exact location of tumor (margins) in pre- or intra-operative use. 

The main research question for this work (and the preliminary work in \cite{BIOIMAGING}) is, how well a classification system learnt from one CLE data set (acquired in the oral cavity) can be applied to another data set (acquired from the vocal folds), and vice versa. Since, in both cases, epithelial tissue is investigated, a robust algorithm should be able to apply knowledge acquired in one domain in the other domain. This would provide a strong hint of generalization towards other locations of the upper aero-digestive tract with similar but not identical epithelia.

\begin{figure} 
\centering
	\includegraphics[height=4cm]{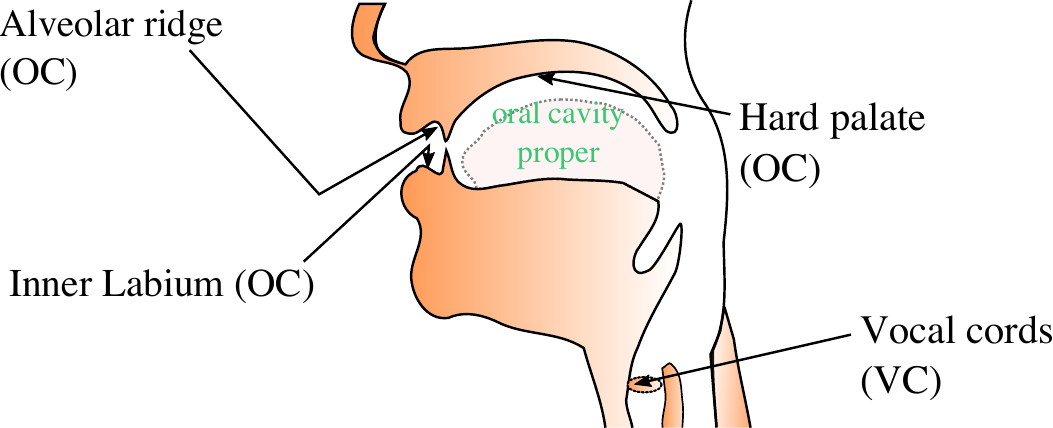}
	\caption{Anatomical locations / examined measurement points from the oral cavity and the upper aero-digestive and respiratory tract. (adapted from \cite{BIOIMAGING})}
	\label{locations}
\end{figure}

\section{Related Work}
Deep learning methods, such as convolutional neural networks (CNN) have recently been used in a variety of image recognition tasks. For the application of cancer recognition, it was shown that, given sufficient training material, they can be used to differentiate various subtypes of skin cancer from photography images \cite{Esteva:2017ct}. In the field of bright light microscopy, CNN-based methods are predicted to become the leading pattern recognition tool \cite{8118310}. 

For the processing and analysis of CLE images, a number of algorithmic approaches have been proposed to date. Our work group was among the first to automatically detect malignancies using manually designed features for classification with support vector machine (SVM) or random forest \cite{Jaremenko:2015kha}. They used a patch extraction from the round pCLE field of view to classify OSCC images from clinically normal epithelium \cite{Jaremenko:2015kha}. Using the same tool chain, Vo \textit{et al.} showed that this detection methodology is also applicable to the detection of SCC of the vocal folds \cite{Vo:2016ux}. On a larger (and thus more realistic) data set of OSCC, the performance of these algorithms decreased, however \cite{Aubreville:2017ux}. We were able to show, that the usage of a CNN for classification of extracted patches greatly improves performance \cite{Aubreville:2017ux}.

One major issue with most medical images is the quantity of data of a specific class for training. For deep networks, the number of parameters increases significantly with every layer, and so does the network's capacity to memorize single instances. An increased quantity of parameters often has a negative effect on generalization, and can lead to over-fitting the training data, in cases where the training data set is limited, which is also true for our (and other authors') CLE data sets. One method to cope with this issue is the reduction of input size of the network, in conjunction with patch extraction from the image and subsequent fusion of separate classification outcomes. This has the additional advantage, that a large portion of the area of the round CLE field of view can be investigated. The amount of patches per image further increases the total batch size for training of the algorithm. 

Another way of approaching this issue was proposed by Murthy \textit{et al.}: They proposed a two-staged cascaded network, where the first stage would only perform a classification with low confidence, and the second stage would only be trained on data that was considered difficult by the first stage \cite{Murthy:2017ky}. For many image recognition tasks, transfer learning from networks pre-trained on large data bases (e.g. ImageNet) has proven to be an effective and well performing approach. Izadyyazdanabadi \textit{et al.} have used transfer learning on CLE images, and have shown that different fine-tuned models each outperformed the same model trained from scratch \cite{Izadyyazdanabadi:2017kp}.

\begin{figure*}
\centering
	\includegraphics[width=\textwidth]{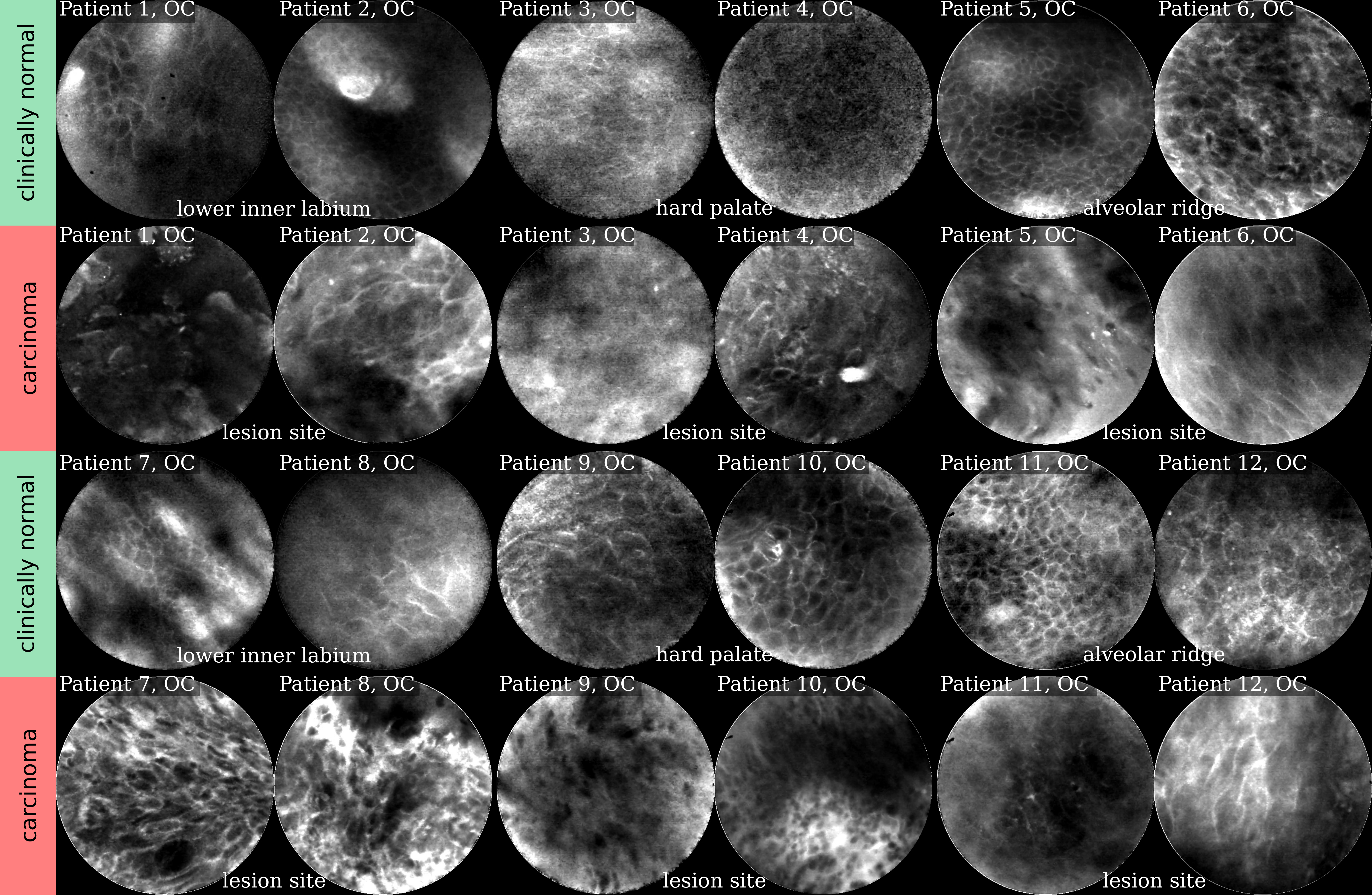}
	\caption{Selected CLE Images acquired from the oral cavity (OC). First and third row show images from clinically normal regions, second and fourth row show images from lesion sites later diagnosed as OSCC.}
	\label{CLE_OC_data}
\end{figure*}

As stated, one problem for classification of pCLE images is the round field of view (see e.g. Fig.~\ref{CLE_OC_data}). Because of this, besides patch-based methods \cite{Jaremenko:2015kha,Vo:2016ux} that extract rectangular patches from the round image area, all image-scale classification approaches known today use squared center crop of the overall image, which reduces the available data for classification \cite{Murthy:2017ky,Aubreville:2017ux,Stoeve:mqZZ9hnc}. 

Since the magnification of pCLE is very high and matching with histology images is almost impossible, image segmentation data of malignancy is hard to derive for the expert, so data sets typically have only image labels. Differentiation of diagnostic vs. non-diagnostic (e.g. artifact-tainted) image parts, however, can be performed by a CLE expert. This sub-image classification can be done supervised (as proposed by Stoeve \textit{et al.} for motion artifacts \cite{Stoeve:mqZZ9hnc}), or also \textit{weakly} supervised (as proposed by Izadyyazdanabadi \textit{et al.}) \cite{Izadyyazdanabadi:2018tm}. Even for the case of malignancy classification, however, a sub-image classification would be interesting for the observer, as it helps interpretation of the image and the classification result.

\section{Material}

For this work we used CLE sequences from four anatomical locations (see Fig.~\ref{locations}): Our original data set consists of images from three clinically normal sites within the oral cavity and, additionally, data from a verified SCC lesion site. Our secondary data set was recorded on epithelium of the vocal cords. 

All images were acquired using a pCLE system (Cellvizio, Mauna Kea Technologies, Paris, France). Written informed consent was obtained from all patients prior to the study. Also prior to the study, approval was granted by the respective institutional review board. Research was carried out in accordance with the Code of Ethics of the World Medical Association (Declaration of Helsinki) and the ethical guidelines of the Friedrich-Alexander-Universit{\"a}t Erlangen-N{\"u}rnberg.

\subsection{Oral Cavity (OC)}

For this study, we included 116 image sequences from 12 subjects with histo-pathologically verified HNSCC in the oral cavity that were recorded at the Department of Oral and Maxillofacial Surgery (University Hospital Erlangen). The study was approved by the IRB of the University of Erlangen-N{\"u}rnberg (reference number: 243\_12 B).

The investigation was conducted chair-side on conscious patients prior to histo-pathological assessment and surgical procedure for removal of the tissue. Imaging was performed on three non-affected sites, i.e. the alveolar ridge, the lower inner labium and the hard palate (see Fig.~ \ref{locations}). Oetter \textit{et al.} conducted a CLE expert recognition assessment based on a subset of the data set used in this work (95 CLE sequences) and found an accuracy of $92.3\,\%$ \cite{Oetter:2016cp}.

\subsection{Vocal Cords (VC)}

The promising results in other anatomical areas motivated the acquisition of CLE images for diagnosis of SCC of the vocal folds, which is the most prevalent form of cancer of the laryngeal tract \cite{Parkin:2005ig}. When first symptoms occur, the primary investigation method is white light endoscopy, which is however known to be insufficient for diagnosis, since a great number of non-malign alterations with similar appearance in endoscopy exists \cite{Goncalves:2017wn}. In the vocal cord area, the motivation for non-invasive micro-structural analysis, as provided by CLE, is even more striking, since extensive sampling of tissue by means of biopsy can cause functional problems such as voice modifications or chronic hoarseness \cite{Cikojevic:2008jd}. Besides its limitations, it was successfully shown that -- given sufficient training -- human experts are able to classify SCC from CLE image sequences successfully, with accuracies between 91.38\% and 96.55\% \cite{Goncalves:wn}. Goncalves \textit{et al.} found a significant training effect (as described by Neumann \cite{Neumann:2011tc}) when judging CLE images on a single still image base. On a similar, different data set with 7 patients, they found 
accuracies by Ear, Nose and Throat (ENT) specialists to be between 58.1\% and 87.1\%, where the non-CLE experienced doctors had a mean accuracy of 67.7\% and those with profound CLE experience one of 82.2\% \cite{Goncalves:2017wn}. 

\begin{figure*}
\centering
	\includegraphics[width=0.83\textwidth]{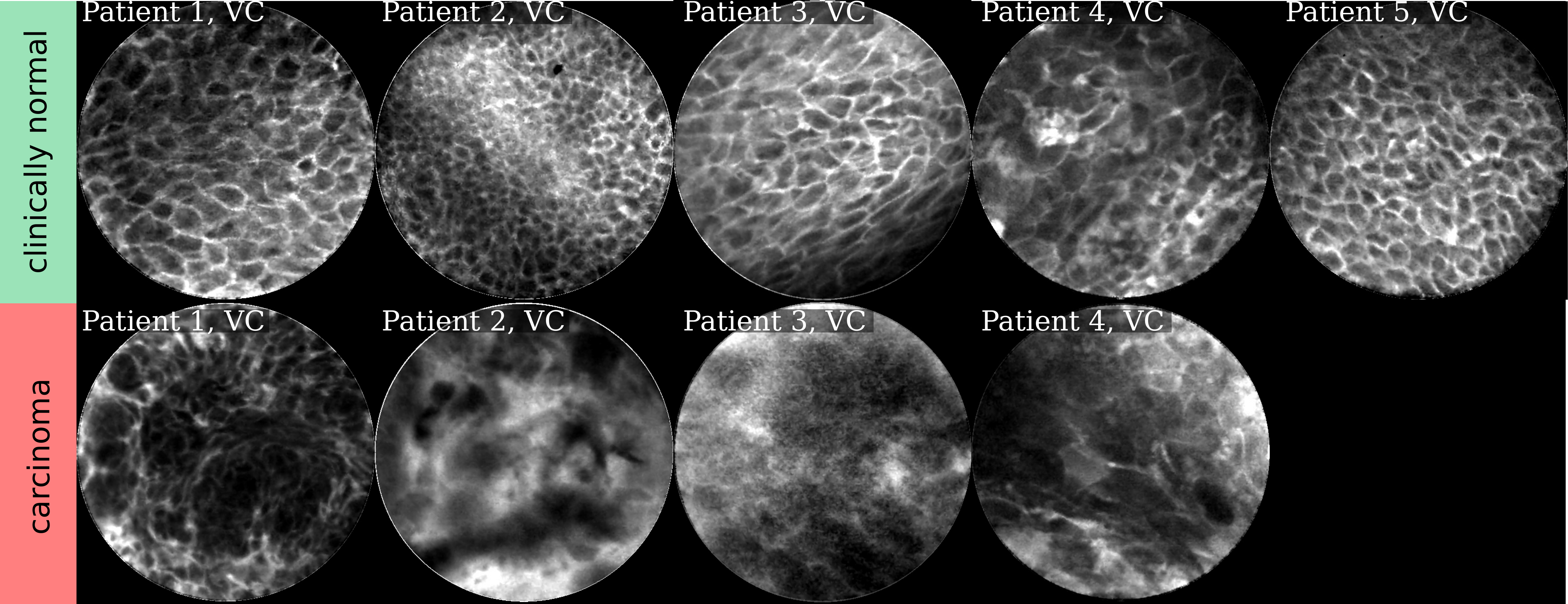}
	\caption{Selected CLE Images acquired from the vocal fold area. In the top row, presumably healthy images are depicted that were taken from the clinically normal contralateral vocal cord of patients with epithelial cancer one of the both vocal cords (shown in the bottom row). Note that for patient 5, only images showing clinically normal tissue were available. }
	\label{CLE_VC_data}
\end{figure*}

For this work, 47 image sequences from five patients were recorded. All patients underwent biopsy subsequent to CLE and were diagnosed with SCC of the vocal folds using the gold-standard of histo-pathology. The images of this data set have been acquired during micro-laryngoscopy at the Department of Otorhinolaryngology, Head and Neck Surgery (University Hospital Erlangen). The study was approved by the IRB of the University of Erlangen-N{\"u}rnberg (reference number: 60\_14 B). For each patient, the contra-lateral vocal cord that was in each case  macroscopically unaffected was also recorded, building a data set representing presumably healthy tissue of the vocal folds. For one patient (patient 5, see Fig.~\ref{CLE_VC_data}), no sequences with histo-pathology-confirmed SCC were available when this data set was created. 

All images were pre-selected by a clinical expert in CLE imaging. This was done to remove diagnostically useless images, e.g. when the probe was not in contact with the tissue. After this cleaning step, the total number of images of this data set is 4,425. Vo \textit{et al.} have done previous investigations on this data set, and found accuracies in grading of between 86.4\% and 89.8\% \cite{Vo:2016ux} using a patch-based method where they found best results for a support vector machine classifier with a feature set based on gray-level co-occurrence matrices \cite{Jaremenko:2015kha}.

\subsection{Image Quality and Artifact Occurrence}
\label{imagequality}
As described by Neumann \textit{et al.} and also Izadyyazdanabadi \textit{et al.}, CLE images are tainted by a range of deteriorations that impede image quality \cite{Neumann:2012ic,Izadyyazdanabadi:2017kp}. A significant part of images within the CLE sequences show low signal to noise ratios, which renders them unusable for either human or algorithmic evaluation. The reasons behind are manifold: the concentration of contrast agent might be insufficient in the tissue under the probe, the contact to the tissue might be lost, or the tissue perfusion with blood containing contrast agent might be too low. All of these factors can rarely be determined in retrospect. We found that low signal to noise ratio seems to be correlated with SCC image sequences. However, for the sake of robustness we assume that this is a non-causal relationship. Thus, very noisy images that were judged to be of no diagnostic use were removed from the data sets after manual inspection.

Comparing the images in Figures \ref{CLE_OC_data} and \ref{CLE_VC_data}, for clinically normal (and presumably healthy) tissue we find a large difference in contrast for some regions. While for images acquired from the alveolar ridge (rows 1 and 3, columns 5\,-\,6 in Fig.~\ref{CLE_OC_data}) and the vocal folds (top row in Fig.~\ref{CLE_VC_data}), contrast is generally good and cell outlines can be clearly spotted for most images, we find a much broader spread in image quality for images acquired at the hard palate and the lower inner labium (rows 1 and 3, columns 1-4 in Fig.~\ref{CLE_OC_data}).  

\begin{figure}[t]
\centering
\includegraphics[width=0.5\textwidth]{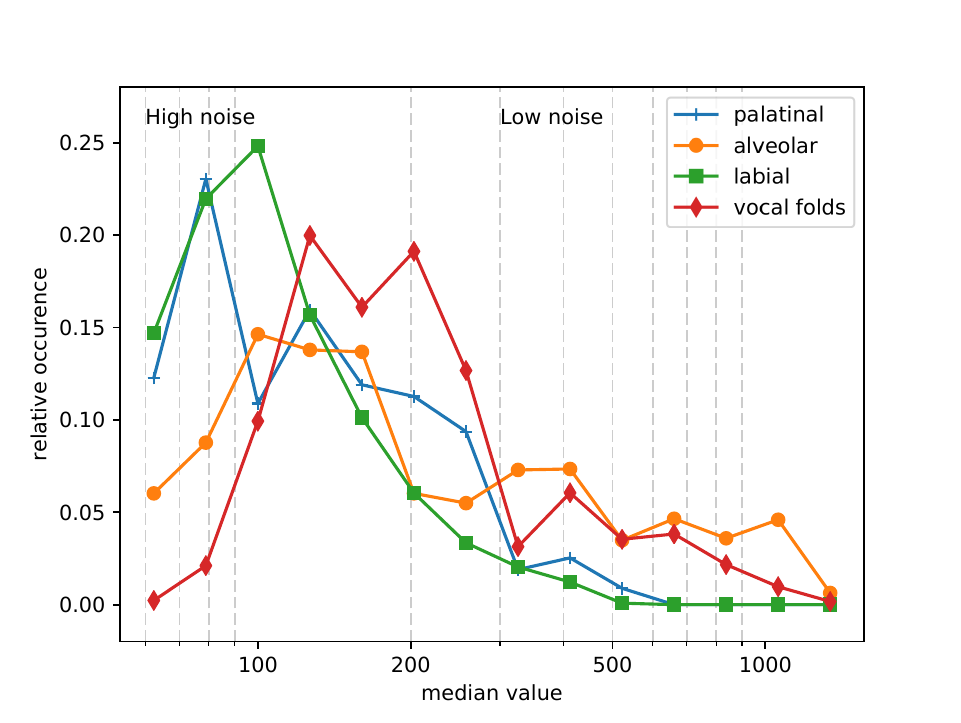}
\caption{Normalized histogram of the median value for the different classes of clinically normal tissue from both data sets. Due to the wide range of pixel values, the histogram is given at log scale. (from: \cite{BIOIMAGING})}
\label{median}	
\end{figure}

In CLE images, the raw pixel value represents the optical (fluorescent) response to the laser light excitation. The optical receiver and analog to digital converter of the CLE scanner in use have a broad dynamic response. Thus, all images presented in this work and also in the CellVizio software are automatically compressed in their value range to fit the 8 bit gray level range of today's screens and to increase contrast for the viewer. The median raw pixel value, however, is indicative of the amount of fluorescent light that was picked up by the scanner, and as such also related to the signal-to-noise-ratio of the image. We use median, as it is more robust to the (typically sparsely distributed) very bright areas representing micro-vessels filled with contrast agent, as e.g. in row 2, column 4 of Fig.~\ref{CLE_OC_data}. We evaluated the median raw pixel value on images for all anatomical regions separately (see Fig.~\ref{median}). For images acquired from the hard palate and the lower inner labium, we find a significantly different distribution than for both other anatomical locations, in that the prevalence of images with low median value (and thus likely low SNR) is much higher for these regions. In contrast, the vocal fold and alveolar ridge area both have a rather low amount of noisy images. This underlines our visual observation derived from the representative images in Fig.~\ref{CLE_OC_data}.

This difference can likely be attributed to the different anatomical properties of the tissue at the respective locations: Epithelium exposed to higher mechanical stress (in our case: usually due to chewing), is known to have a higher degree of cornification \cite{Rohen:2000ut}. The vocal cords, due to not being exposed to mechanical stress, consist of multiple layers of uncornified squamous epithelium\cite{Rohen:1994tv}. The alveolar ridge, located in the vestibule and outside of the \textit{oral cavity proper} (where food is grinded and predigested), is normally not subject to high levels of mechanical stress. The hard palate, in contrast,  is part of the chewing process and known to have a high degree of cornification \cite{LullmannRauch:DZUvi2yA}. The inner lip (labium), also located in the vestibule, is generally not considered a cornified epithelium. The images in our oral cavity data set, however, were taken at the intersection of mucous epithelial tissue in the vestibule and the skin-covered outer lip, which has an epidermal layer which is associated with a high level of cornification \cite{Rohen:1994tv}.

\section{Methods}

\begin{figure}
\centering
	\includegraphics[width=\textwidth]{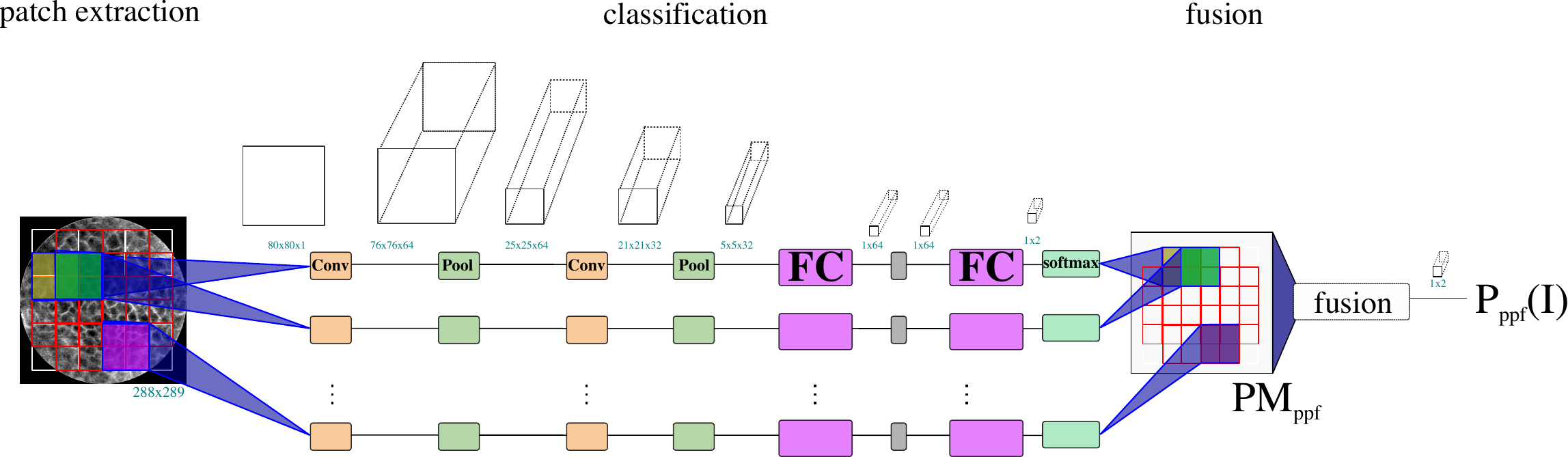}	
	\caption{Overview of the patch probability fusion (ppf) method \cite{Aubreville:2017ux}. Patches are extracted from the image, classified by a DNN using two layers of convolutional, pooling and finally two fully connected (FC) layers and subsequently probabilities are fused to a per-image probability.}
	\label{overview_ppf}
\end{figure}

In this paper, we want to compare two methods of detection that differ significantly in their structure. Firstly, we assess a patch-based method using a CNN for individual classification of the patches (see Fig.~\ref{overview_ppf} and \cite{Aubreville:2017ux} for further details). The patch classification is projected on the originating image coordinates, thereby deriving a probability map $\mathrm{PM}_\mathrm{ppf}$. Subsequently, the maps are fused to a single scalar for each class, denoted as image probability $\mathrm{P}_\mathrm{ppf}$. We denote this approach \textit{patch probability fusion} (ppf). The approach is subject to the supposition that image labels can always be assumed to be also correct for individual patch labels, which, in our experience, holds true for a vast majority of images in our data set, since rarely the actual margin of a tumor is shown within an image with normal tissue structure in one part and malignant structural changes in the other.

Secondly, we perform a classification on image level, where a preprocessed complete image is fed through a deep network based on the stem of an Inception V.3 network \cite{Szegedy:2014tb} which will be described in detail in section \ref{wslmethod}.

\subsection{Experimental Setup}
We performed five data experiments for this study: 

\subsubsection{Algorithmic generalization}
One major question is, how well the algorithms work on both data sets, when trained and evaluated on the same data set (using leave-1-patient-out (LOPO) cross-validation). This experiment aims to give us hints about algorithmic robustness.
\begin{enumerate}
	\item[1] (OC) Training on 11 patients of the OC data set, test on 1 patient (LOPO cross-validation).
	\item[2] (VC) Training on 4 patients of the VC data set, test on 1 patient (LOPO cross-validation).
\end{enumerate}
The results given for these experiments represent an evaluation on the concatenated result vector of all cross-validation steps.

\subsubsection{Medical data generalization}
Since in both data sets, epithelium was scanned using the same technology, we assumed that classification knowledge acquired from one domain could generalize to the other domain.
\begin{enumerate}
	\item[3] (OC/VC) Training on all patients of the OC data set, test on all patients of the VC data set.
	\item[4] (VC/OC) Training on all patients of the VC data set, test on all patients of the OC data set.
\end{enumerate}

\noindent Finally, we concatenated both data sets and performed yet another LOPO cross-validation on the joint data set.
\begin{enumerate}
	\item[5] (OC+VC) Training on 16 patients of the concatenated data set, test on 1 remaining patient (LOPO cross-validation).
\end{enumerate}

\subsection{Training of the Patch Classifier}

We trained the patch classification network for a fixed number of 60 epochs, using an 3-fold augmented data set with random rotations of multiples of $\pi/2$ on patch level. All network weights have been initialized randomly. We used cross-entropy loss with the ADAM optimizer (learning rate: $10^{-2}$). 

\subsection{Whole Image Classification with Activation Maps}
\label{wslmethod}
Especially in the medical domain, two important requirement towards examination methodology are predictability and transparency. This contrasts with deep networks, essentially representing a complex function with unknown content and trained numerically. Therefore, insight into the method is difficult to attain. Especially since for domains where image quantity is low, overfitting plays an important role, medical experts and authorities may demand insights into intermediate classification results. As such, the aforementioned patch probability fusion (ppf) approach gives insights into classification results of different areas of the image, but its receptive field is also limited to the patches it classifies. The human expert, however, will always assess the whole image, and consequently also the labels are (for our data set) given only on image level. 

\begin{figure}
\centering
	\includegraphics[width=\textwidth]{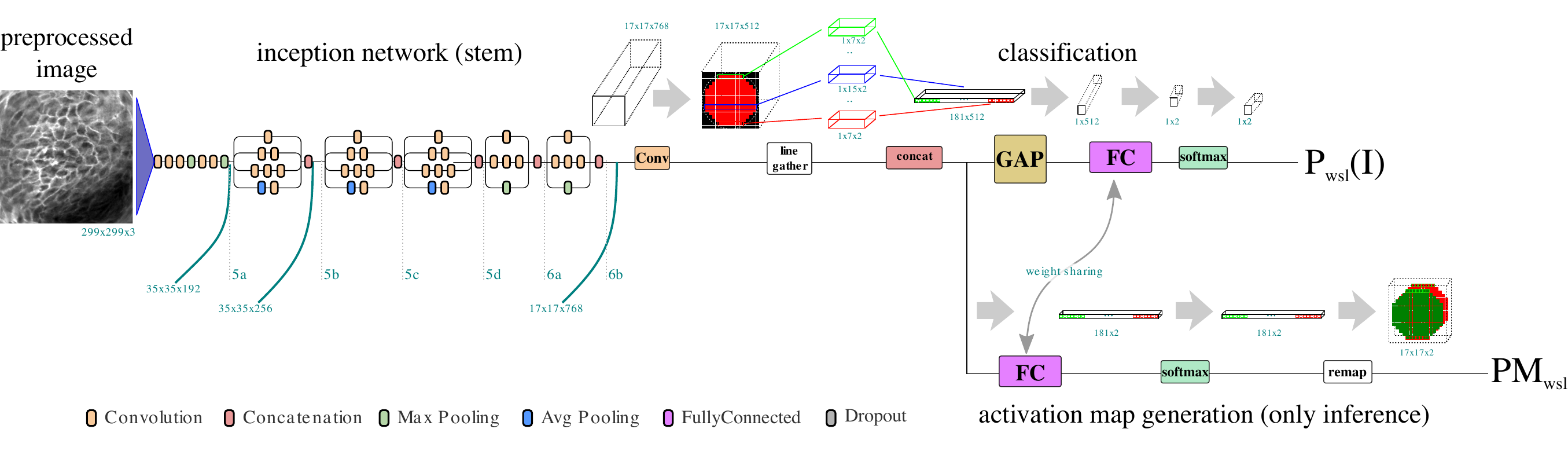}	
	\caption{Overview of the proposed whole image classification approach. Preprocessed images are fed through the stem of a pre-trained Inception V.3 network, then a convolutional (Conv) layer, a global average pooling (GAP) operation and finally a fully connected (FC) layer are attached. During inference, a class activation map is derived.}
	\label{overview_wsl}
\end{figure}

One hint towards understanding image-level classification was given by Oquab \textit{et al.}\cite{Oquab:2015fa}, a method that is commonly denoted as \textit{weakly-supervised} learning. Their idea was to use a convolutional neural network that consists only of convolutional layers and pooling / striding operations. This leaves intact the localization of the image. Next, a global pooling operation is employed, resulting in an $1 \times 1 \times C$ image, where $C$ is the number of channels/filters of the previous layer. Then, a final fully connected layer creates the mapping towards the output classes (see classification branch in Fig.~\ref{overview_wsl}). This approach has some nice properties: Since all operators are agnostic of the position on the actual image and all image regions are included into the final classification with equal weight, it is ensured that similar structures are recognized equally and a certain translational robustness can thus be expected. During inference, a second branch is added to the network prior to the average pooling layer, which maps the activations of said layer to class activity maps for each class, that can be easily visualized and gives insights into possible interpretations of the recognition. 

Weakly supervised classification is especially useful, if only image-labels are available, but information about the subparts of the image would be informative. In our database of CLE images, we have these image-level class labels, as no further annotation was made available or is feasible to perform, since structures acquired during CLE image acquisition are hard to be identified exactly in the H\&E-stained histology images that were taken as diagnostic samples from the same area.

For the case of CLE the round circle of view is a challenge for image-level algorithms, due to steep edges on the corner. To circumvent this, we employ a preprocessing scheme described in section \ref{preprocessing} which ensures similar statistical probabilities across the whole image. 

Still, since the newly created border area does not contain useful information, we aim to limit the networks attention to the area inside of the circle. For this, we suggest to use masking operations within the network, preceding the global average pooling (GAP) layer. The masking operation cuts position-dependent fragments out of the original image, that are due to the nature of being located just before the GAP layer, however, no longer dependent on the actual position. 

We denote the respective previous layer of a network to be $\mathbf{U}$ with its elements $u_{i,j,c}$. The average pooling operation $F_\textrm{GAP}: \mathbf{U} \rightarrow \mathbf{V}$, with $\mathbf{U} \in \mathbb{R}^{W \times H \times C}, \mathbf{V} \in \mathbb{R}^{1 \times 1 \times C}$ and $W \times H$ being the spatial resolution of the network and $v_C$ being the elements of $\mathbf{V}$ is defined as:

\begin{equation}
 v_c = \frac{1}{W\cdot H}\sum_j^W \sum_i^H u_c(i,j)
\end{equation}

In order to restrict the attention of the algorithm to areas that have a valid image, we apply a masked average pooling operation $F_\textrm{MAP}: \mathbf{U} \rightarrow \mathbf{V}'$ with $\mathbf{V}' \in \mathbb{R}^{1 \times 1 \times C}$ the elements of the resulting vector $v'_c$ as:

\begin{equation}
v^*_c = \frac{1}{\sum_i^W \sum_j^H \delta(i,j)}\sum_j^W \sum_i^H \delta(i,j) \cdot u_c(i,j) 
\end{equation}
with:
\begin{equation}
\delta(i,j) =  \sigma \left(r^2 - \left(j-\frac{H}{2}\right)^2 - \left(i-\frac{W}{2}\right)^2 \right)
\end{equation}

where $r$ is a constraining radius of the field of view, and $\sigma(x)$ is the step function. 

We attach the aforementioned masked global average pooling operation at an intermediate endpoint of Szegedy's Inception V.3 model \cite{Szegedy:2014tb}, denoted as \textit{6b} (see Fig.~\ref{overview_wsl}). In previous experiments, we found this layer to be a good compromise between complexity, generalization and performance. After the masked GAP operation, we attach a fully connected layer to retrieve a mapping to two output classes. Finally, a softmax operator is used to derive probabilities on image level. The network is trained using binary cross-entropy as loss function.

During inference, a second branch (entitled 'activation map generation' in Fig.~\ref{overview_wsl}) is added just before the GAP operation, starting with a fully connected layer with shared weights to the fully connected layer in the classification branch. After the softmax, a class activation map of size $17\times17$ can be retrieved. 

\subsection{Preprocessing of the Image Classifier}
\label{preprocessing}
To avoid problems in optimization due to the steep edges of the round field of view of pCLE images, we performed a circular extrapolation, as described in the following steps:

\begin{itemize}
	\item Transformation of the image using a linear to polar transformation.
	\item Concatenation of the image in polar representation with its flipped version (along the distance axis). 
	\item Transformation into the linear domain.
	\item Cropping of the square representing the original image coordinates.
\end{itemize}

\noindent This creates a mirroring around the circle that is defined using the radius in the linear to polar transformation. An example preprocessed image can be seen in Fig.~\ref{overview_wsl}. 

\subsection{Training of the Whole Image Classifier}

Using Inception V.3 \cite{Szegedy:2014tb} as its base topology, the network capacity of the image classifier is significantly higher than that of the patch classifier. To leverage transfer learning, we initialize this network stem with pre-trained values acquired on the ImageNet database. With this, we try to benefit from the knowledge about shapes that can be derived from the real-world images used for pre-training this stem. We assume that this knowledge generalizes better than any conclusions that the model would derive from CLE data, so we train the stem with a reduced learning rate ($10^{-4}$ vs. $10^{-6}$). 

In order to avoid overfitting, we make use of an early stopping mechanism. For this, two patients of the training data set (for the VC data set: 1 patient, due to its smaller size) are used as validation set. The training is initially run for 2 epochs. After the third and every following epoch, the classifier is evaluated for performance on the validation data set. In case of performance decrease, coefficients from the last epoch are restored and training is stopped. We allow for a maximum of 10 epochs, which was in practice never reached in our experiments. We assume that a validation on a different patient should allow to draw conclusions about overfitting within the same data collective. Further, we arbitrary rotation (before the preprocessing) as augmentation during training. 

\section{Results}

\subsection*{Overall Results}

Both algorithmic approaches seem to be applicable to both data sets with overall comparable results. The image-based classification yields better results for the oral cavity data set (AUC of $0.9687$ vs. $0.9550$, see Table \ref{results}), while the overall results are similar for the vocal cords data set (AUCs of $0.9484$ vs. $0.9550$). 

For the medical data generalization task, which was the main research question in this work, we find a good performance for both algorithms when trained on the OC data set and applied on the VC data set. In contrast, the generalization from vocal folds to oral cavity performs significantly worse, and even more so for the whole image classification method.

Not surprisingly, we find that the overall performance of both approaches benefits from the larger (joint) data set in the OC+VC experiment. 

\begin{table*}[ht!]
\centering
	\begin{tabular*}{\textwidth}{|l@{\extracolsep{\fill}}|c|c|c|c|c|c||c|c|}
	\hline
	\multirow{2}{*}{Condition} & 
		\multicolumn{2}{c|}{Accuracy (\%)} &
		\multicolumn{2}{c|}{Precision (\%)} &
		\multicolumn{2}{c||}{Recall (\%)} &
		\multicolumn{2}{c|}{ROC AUC (\%)} \\
		& ppf & image & ppf & image & ppf & image & ppf & image \\
%		Condition & Accuracy & Precision & Recall & ROC AUC \\
		\hline
		OC    & 88.34\% \cite{Aubreville:2017ux}& 91.63 \% & 85.40\%   & 91.06\% & 91.10\% & 91.43\% & 0.9550 & \textbf{0.9687} \\
		VC    & 91.39 \% & 89.97\% & 93.64\% &93.66\% & 92.03\% & 89.50\% & 0.9484 & \textbf{0.9550}\\
		OC/VC & 89.45 \% & 86.33\% & 87.47\% &82.33\% & 96.37\% & 98.58\% & 0.9548 & \textbf{0.9644}\\
		VC/OC & 68.53\% & 61.68\% & 60.81\% &55.60\% & 95.63\% & 97.43\% & \textbf{0.8484} & 0.7694 \\
		OC+VC & 90.81\% & 92.63\% & 90.12\% &92.89\% & 92.59\% & 93.04\% & 0.9697 & \textbf{0.9762} \\ 
		\hline
	\end{tabular*}
	\caption{Results of all image and patch based (ppf) tests. For the cross validation cases OC, VC and OC+VC, the results were calculated on the concatenated result vector of all cross validation steps.}
	\label{results}
\end{table*}

\begin{figure}
\centering
\includegraphics[width=0.5\textwidth]{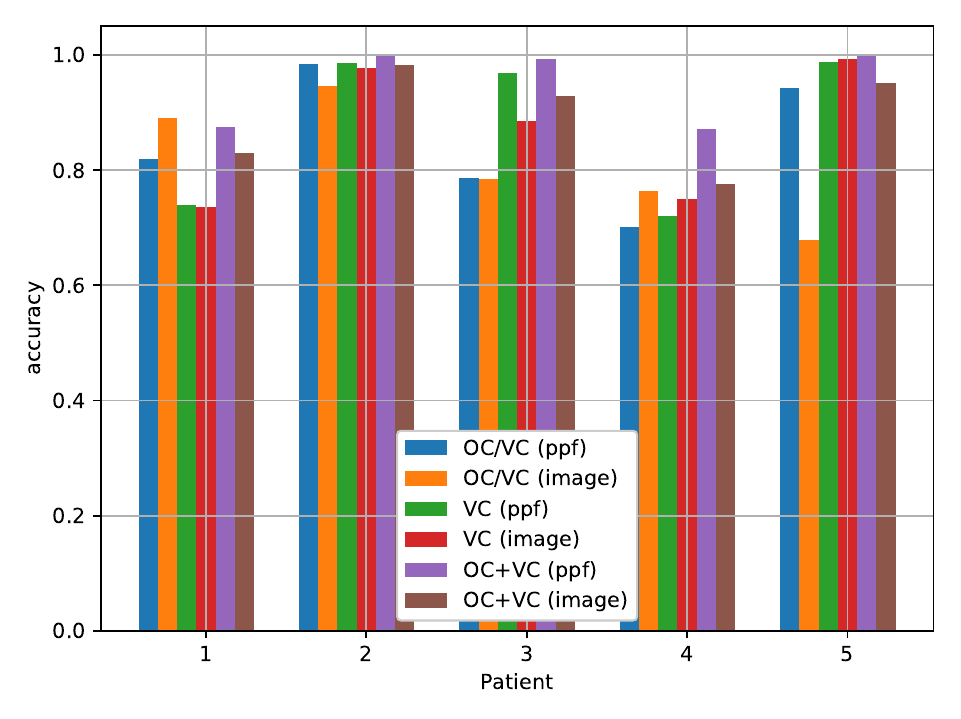}
\caption{Accuracy for all patients with both classes of the vocal fold data set.}	
\label{acc_vc}
\end{figure}

\begin{figure}
\centering
\includegraphics[width=\textwidth]{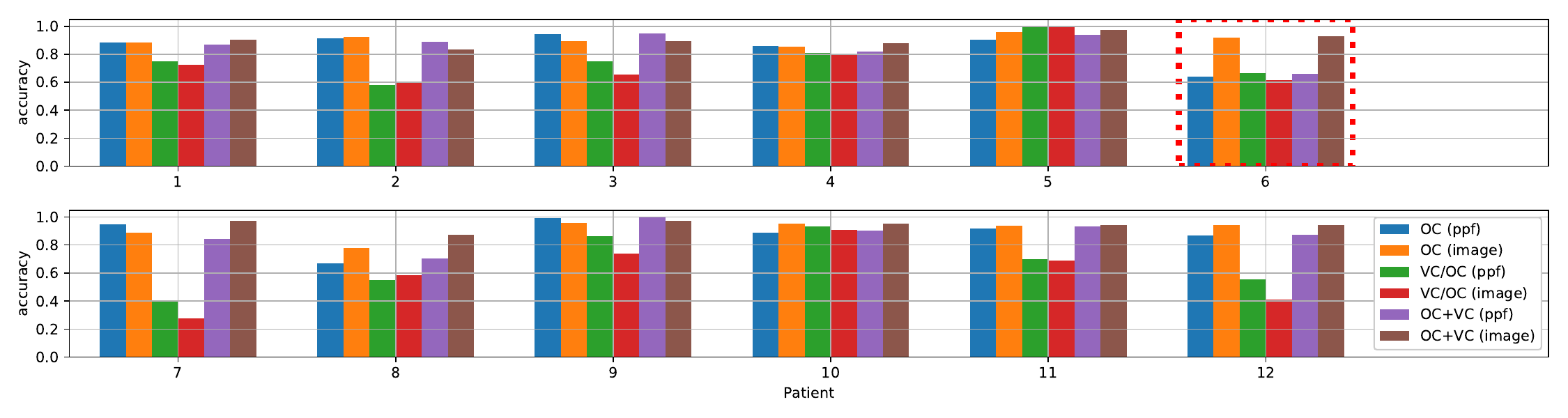}
\caption{Accuracy for all patients of the oral cavity data set. While for most patients, image level recognition and ppf method have similar results, patient 6 shows significantly improved results for the proposed approach.}	
\label{acc_oc}
\end{figure}

\subsection*{Single Patient Results}

Investigating single patient results, we find significant differences between individual patients. Where overall accuracies of some patients were approaching perfect recognition, others showed mediocre performance -- this is most striking in the VC data set (see Fig.~\ref{acc_vc}).

\begin{figure}
\centering
\subfigure[Histogram of probabilities. Top: Proposed image classification method, bottom: patch-based method]{\includegraphics[height=3.8cm]{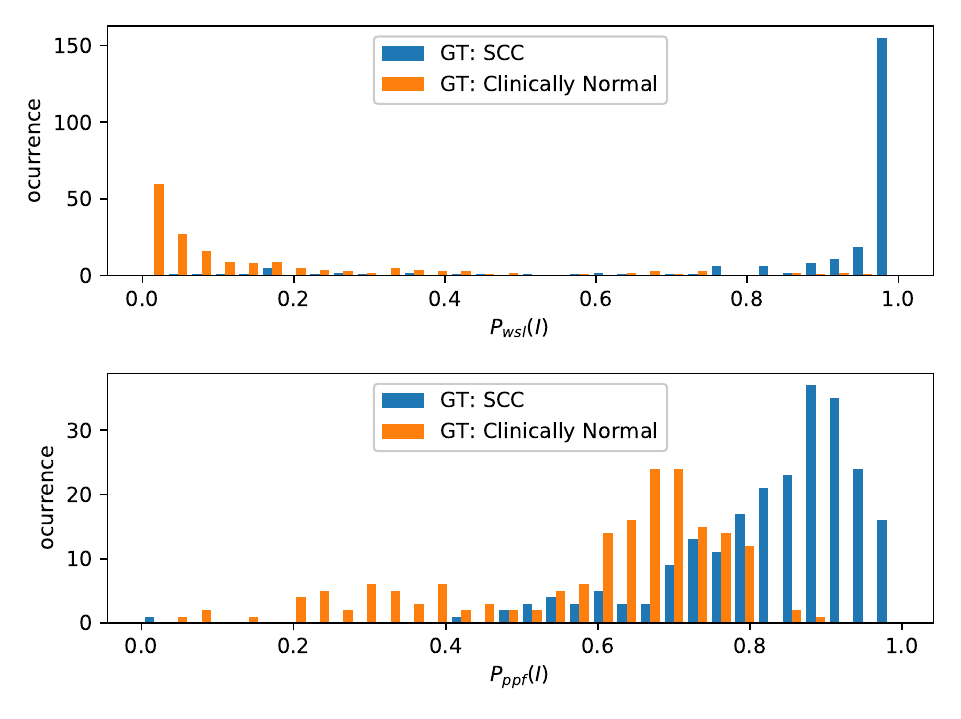}}
\hspace{0.3cm}
\subfigure[Example original images (left) with class map of the proposed image classification method (middle) and the ppf method (right).]{\includegraphics[height=3.8cm]{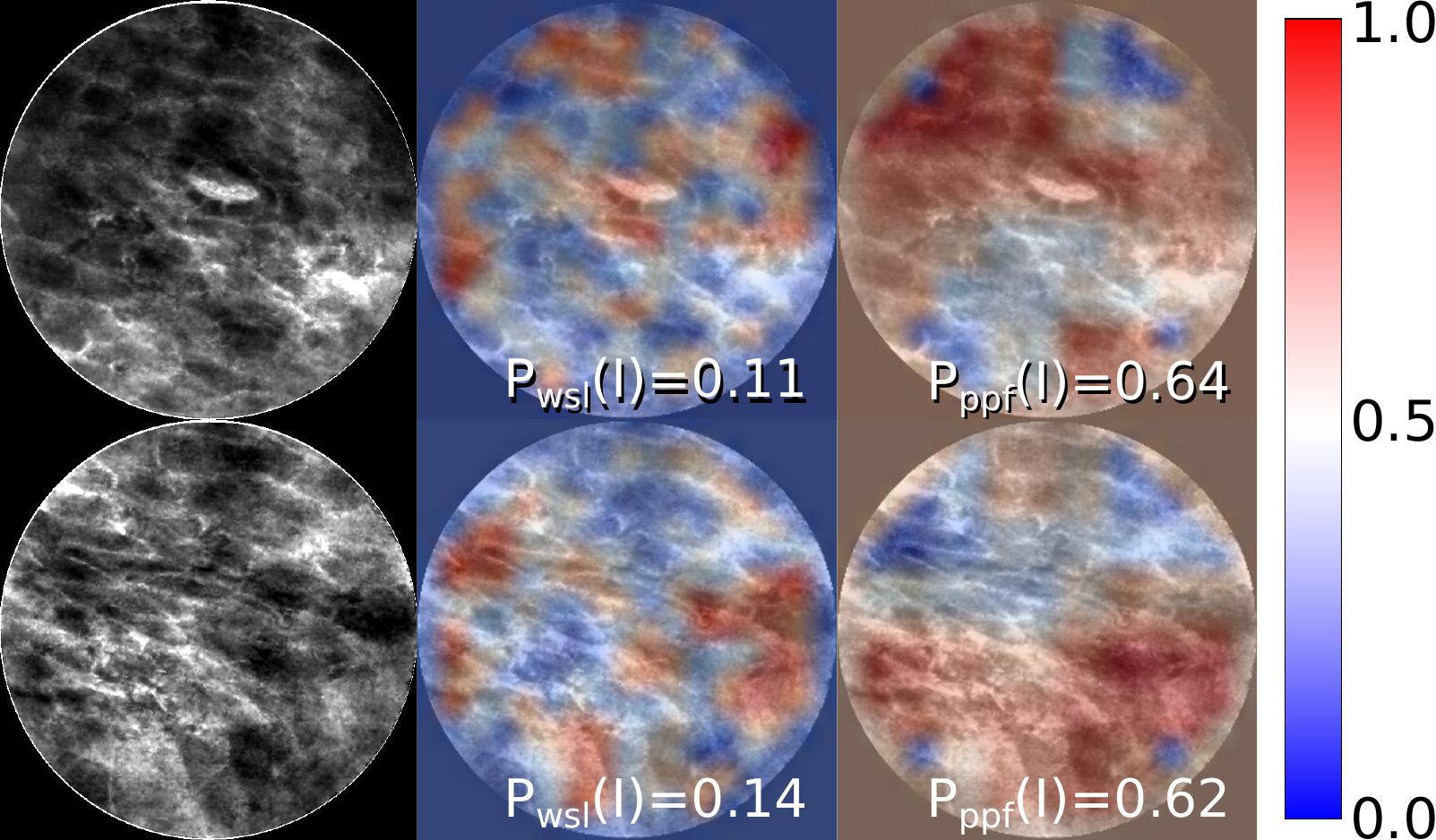}\label{cmpatient6}} 
\caption{Comparison of results for both algorithmic approaches for patient 6 of the OC data set. As the histogram indicates, a high number of false positives is produced by the patch-based approach, leading to a decreased accuracy (65.93\% vs. 91.91\%).}
\label{patient6}
\end{figure}

We further find that classification results on one particular patient in cross validation (patient 6, see Fig.~\ref{acc_oc}) benefit significantly from the whole image approach in both, the OC and the OC+VC experiment. This seems to be of strong influence on the overall increased performance for this condition. In the image probabilities produced by both approaches for this patient, we find that the ppf approach produces a significant amount of false positives (Fig.~\ref{patient6}, $P_{ppf}>0.5$ in clinically normal images), resulting in a low accuracy of $65.93\%$. While this might seem like a training anomaly at first, it is present in both, the concatenated OC+VC data set as well as the OC data set alone. In contrast, for the whole image recognition method, the accuracy for this patient is $91.91\%$.

\section{Discussion}

The results indicate that the recognition worked overall well for both methods. Both methods were applicable on both data sets, with slight advantages for the image level recognition proposed in this work. The prediction from VC data set to the OC data set, however, did not lead to satisfactory results.  While differences in data set size might account for some loss of generalization, it is presumably not the only reason that the classifier trained on the VC data set shows significantly decreased performance on the OC data set. The much lower precision (see Table \ref{results}) in this experiment suggests the interpretation, that the underlying image material also differs significantly, which is backed up by our visual and statistical analysis from section \ref{imagequality}. Further, the anatomical differences (especially w.r.t cornification of tissue) influence these images in an explicable manner. With this, we would expect that a classification system trained on uncornified epithelial tissue would likely not be applicable directly to cornified epithelial tissue. Since the OC data set contains both tissue manifestations, the observed generalization (cf. Table \ref{results})  to the VC data set that contains only uncornified epithelium appears likely.

While images acquired from patient 6 (as depicted in Fig.~\ref{patient6}) are certainly not showing perfectly ordered epithelial tissue, they represent a physiological variation. In this case, the whole image approach with its much wider algorithmic receptive field can make use of the overall image information and classify most images correctly. The distribution also shows a difference between a final softmax (as in the image level classification) and an averaging over different softmax outputs: the amount of samples with low confidence scores is extremely limited in the upper plot, while the averaging by the ppf method in the lower plot generates a more widespread distribution. The color-coded class map in Fig.~\ref{cmpatient6} also indicates that, while certain low-contrast parts of the image are assigned higher class probabilities for carcinoma in the whole image approach, they do not influence the overall probability (color coded in the background of the image).

Furthermore, as both algorithmic approaches profit from an enhanced data set, it is obvious that the limited size of our current data set cannot provide a final answer to the question of algorithmic generalization. The whole image approach introduced in this paper, however, has the much higher algorithmic capacity and should thus be able to adjust to a wider range of physiological alterations, as could be expected when the data set further increases.

While our approach is related to weakly supervised image segmentation, it is important to state that we did not aim to segment the image between \textit{cancer} and \textit{healthy} regions. The reason to use this architecture was rather to give more insights into image cues, used by the method for the image classification. This shows also the close relation between both methods compared in this work: Both perform averaging of some kind of class maps on the complete image, while the GAP-based global feature averaging can encode a much broader scope on image level, as the final mapping to the output classes is only performed with the final fully connected layer. It is worth pointing out that without this layer, both approaches are similar (if we ignore the different filter layers). While patch based approaches usually have a higher flexibility in handling (such as, e.g. for class distribution equalization during training), the image-based approaches with an average pooling layer have a higher computational performance.

A major limitation of our study is that we do not have histo-pathological proof for any of our clinically normal CLE images, since extraction of tissue from clinically normal regions would be ethically questionable, and histo-pathology confirmed squamous cell carcinoma for all patients in our data set. However, we are confident that a larger data set that our research group is preparing to record will also include verified healthy tissue. 

It must be stated, however, that it is questionable if a negative diagnosis for SCC can be retrieved using CLE alone. CLE has a limited penetration depth, and thus tumors that spread within the submucosa could be invisible for CLE. A combination with other methods, such as Raman spectroscopy \cite{Knipfer:2014jm} or Optical Coherence Tomography \cite{Betz:2015ge} could yield improved results. 

The results of this study imply that our method could generalize well for all kinds of epithelial cancer diagnosis where CLE is applicable. This would, however, obviously be a too broad statement given the limited amount of patient data available for this work. Our future work will thus focus on the acquisition of a larger and more diverse data set. The fact that the concatenated data set improved performance for both domains indicates that an enlarged data set will improve performance as well as robustness.

\section{Summary}

In this work we evaluated two approaches for automatic classification of confocal laser endomicroscopy images for the detection of head and neck squamous cell carcinoma, with a focus on the ability to generalize knowledge learnt from one anatomical location (oral cavity) to another (vocal folds). 

The patch-based classification approach and the newly introduced whole image classification system both show good performance when evaluated on the same entity. The image-based classification system showed better performance (accuracy of $91.63\%$ vs. $88.34\%$) on one of the data sets, whereas results on the other data set were similar.

The generalization tasks, where a classification system learnt from CLE images of the oral cavity was applied on vocal fold CLE images, showed promising results. Concatenation of both data sets led to an overall improved accuracy of $92.63\%$ with the image classification system. We expect that a further increase of data amount will significantly improve performance on this recognition task.

%%%%% References %%%%%

\bibliography{literature}   % bibliography data in report.bib
\bibliographystyle{ieeetr}   % makes bibtex use spiejour.bst

%%%%% Biographies of authors %%%%%

\end{spacing}
\end{document}